\documentclass{article}
\usepackage{ijcai26}

\usepackage{times}
\usepackage{soul}
\usepackage{url}
\usepackage[hidelinks]{hyperref}
\usepackage[utf8]{inputenc}
\usepackage[small]{caption}
\usepackage{graphicx}
\usepackage{amsmath}
\usepackage{amsthm}
\usepackage{booktabs}
\usepackage{algorithm}
\usepackage{algorithmic}
\usepackage[switch]{lineno}

\usepackage{subcaption}

\usepackage{amssymb}

\title{
A Survey of Personalized Federated Foundation Models for \\ Privacy-Preserving Recommendation\thanks{Accepted at IJCAI--ECAI 2026 (Survey Track).}
}

\author{
Zhiwei Li$^1$ 
\and
Guodong Long$^1$\and
Chunxu Zhang$^2$\and
Honglei Zhang$^3$\and
Chengqi Zhang$^4$\and
Jing Jiang$^1$\\
\affiliations
$^1$ Australian Artificial Intelligence Institute, University of Technology Sydney\\
$^2$ College of Computer Science and Technology, Jilin University \\
$^3$ School of Computer Science and Technology, Beijing Jiaotong University \\
$^4$ Department of Data Science and Artificial Intelligence, The Hong Kong Polytechnic University \\
\emails
\texttt{zhw.li@outlook.com},
\texttt{guodong.long@uts.edu.au},
}

\begin{document}

\maketitle

\begin{abstract}
Integrating Foundation Models (FMs) into recommendation systems is an emerging and promising research direction. 
However, centralized paradigms face growing pressure from privacy concerns and strict regulatory requirements. 
Federated learning offers a viable solution that enables collaborative model refinement while keeping raw user data on local devices or organizational silos. 
Yet, applying FMs in this setting creates a fundamental tension, where the system must balance the leverage of global knowledge with the necessity of capturing user personality.
This survey provides a comprehensive overview of Personalized Federated Foundation Models for privacy-preserving recommendation, and reviews recent progress in this emerging field.
We first analyze personalization techniques that function effectively under federated settings. 
Furthermore, we discuss the adaptation of foundation models to such federated architectures to balance generalization with user-specific needs for achieving privacy-preserving recommendation. 
In contrast to existing reviews, our work specifically emphasizes the architectural intersection of federation, personalization, and foundation models. \looseness=-1
\end{abstract}

\section{Introduction}
\label{sec:intro}

Recommendation services constitute a fundamental component of the modern digital ecosystem, and rely on aggregating fine-grained interaction data to infer user preferences~\cite{ko2022survey}.
However, this centralized paradigm faces severe systemic constraints with intensifying imperatives for data sovereignty and privacy.
Regulatory frameworks like GDPR~\cite{voigt2017eu} enforce strict constraints on data usage. 
Simultaneously, industrial developments highlight the friction between global service delivery and local data governance. 
Notable instances, such as the scrutiny surrounding TikTok~\cite{aharazi2024contextualizing}, underscore the principle of data sovereignty, which requires user information to remain physically stored within national jurisdictions.
In addition, Australia's Social Media Minimum Age framework~\cite{AusSMMAAct2024} mandates rigorous controls over processing data from minors.
These pressures push the industry to seek alternatives that expand beyond their home markets into a global environment, decoupling service quality from centralized data location. \looseness=-1

\begin{figure}[t]
    \centering
    \includegraphics[width=1\linewidth]{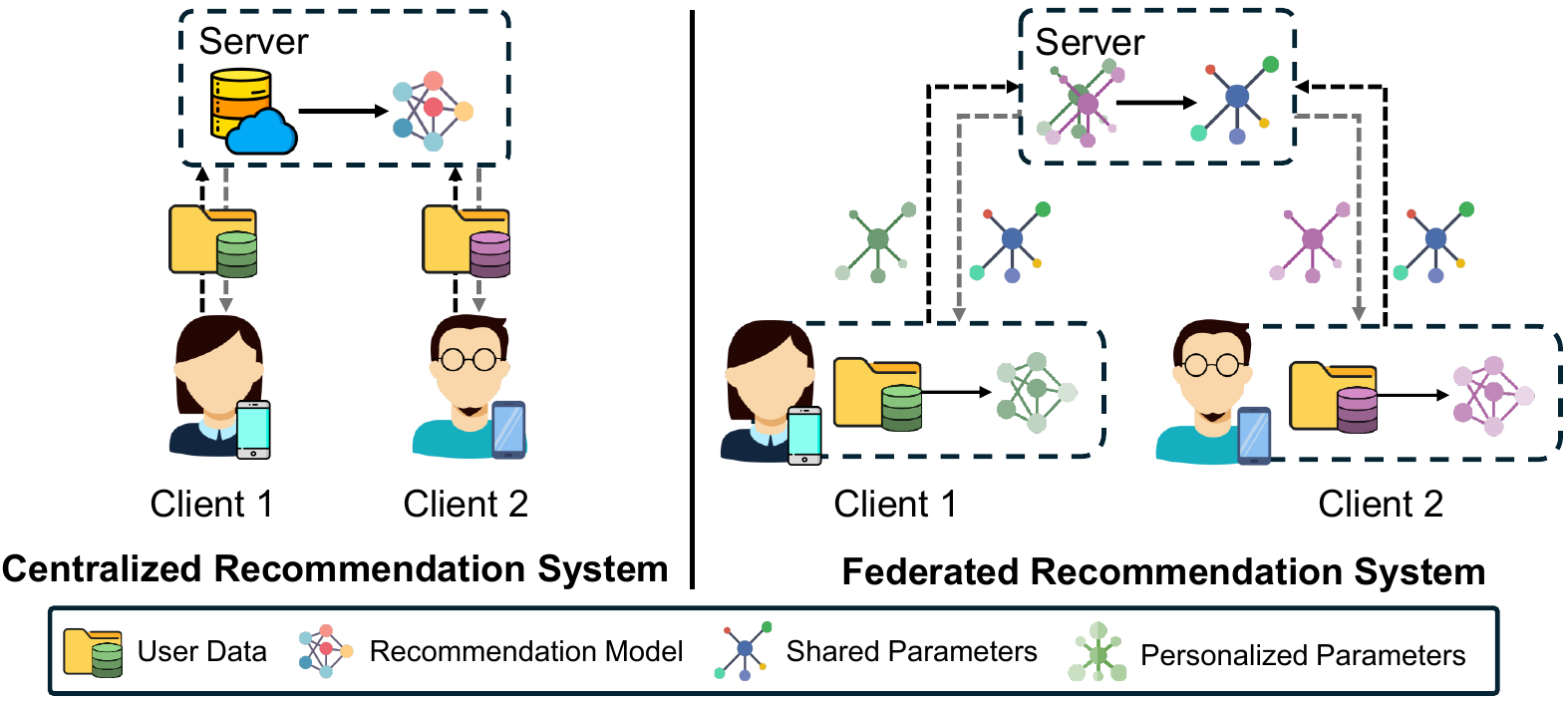}
    \caption{Comparison of Centralized RS (Left), where all user data is server collected, versus FRS (Right), where raw user data and local models remain on client devices, preserving data locality.}
    \label{fig:CRS_vs_FRS}
\end{figure}

To mitigate these risks, Federated Learning (FL)~\cite{mcmahan2017communication} offers a distributed paradigm to establish a privacy-aware and regulation-compliant framework for recommendation at scale. 
Applying this paradigm to Recommendation Systems (RSs) enables data localization with a decentralized service architecture, leading to Federated Recommendation Systems (FRSs).
The conceptual framework is illustrated in Fig.~\ref{fig:CRS_vs_FRS}. 
Unlike centralized RSs which train the models on aggregated logs collected by the platform, FRSs force each client to train a local model using the user's private data, thereby maintaining data locality~\cite{sun2024survey}.
In FRSs, the central server functions solely as a coordinator to aggregate model updates, such as gradients, rather than collecting raw data.
FRSs thus offer a viable pathway to navigate the tension between data utilization and regulatory compliance at scale. \looseness=-1

While FRSs solve the issue of data locality for privacy, modern recommendations require deep semantic understanding.
Foundation Models (FMs) demonstrate remarkable capabilities in discerning complex patterns and forging transferable representations from extensive datasets~\cite{zhou2025comprehensive}, which is particularly useful when client data are sparse or heterogeneous.
FMs generalize well from sparse data and adapt efficiently to downstream tasks~\cite{zhang2024scaling}, thereby acting as a robust semantic backbone of RSs.
Integrating FMs into the federated infrastructure leads to Federated Foundation Models (FFMs)~\cite{yu2023federated}.
FFMs leverage pre-trained knowledge to enhance local learning within the decentralized framework, embodying a distinct approach to collaborative intelligence across heterogeneous clients~\cite{ren2025advances}. \looseness=-1

However, simply deploying FFMs is insufficient for recommendation tasks.
Recommendations demand handling highly idiosyncratic user traits, not just generic semantic knowledge. 
A single global FFM cannot capture the diverse and distinct personalities of all users.
This necessitates Personalized Federated Foundation Models (PFFMs) for privacy-preserving recommendations. 
PFFMs must architecturally decouple shared global knowledge from unique individual characteristics.
Effective PFFMs must therefore actively \textit{learn to preserve} user personality.
This involves treating user traits as sensitive semantic structures to be isolated within the local adaptation process.
This survey focuses on this critical intersection of personalization, federation, and foundation models for privacy-preserving recommendation. \looseness=-1

To our knowledge, this is the first survey dedicated to Personalized Federated Foundation Models for recommendations, and makes the following contributions to the field:
\begin{itemize}
\item We formally define the scope of this emerging area.
We establish a novel perspective centered on the active preservation of user personality semantics, rather than passive data protection for recommendation.
\item We formulate a principled framework for achieving this preservation, and detail the necessary mechanisms to architecturally decouple generalized knowledge from individualized adaptations within federated systems.
\item We conduct a principled analysis of fundamental challenges, and dissect theoretical conflicts between gradient-based utility optimization and geometric constraints necessary for semantic privacy boundaries.
\item We outline critical future directions, and propose new paradigms to bridge current gaps in modeling stable user traits under strict privacy conditions.
\end{itemize}

\section{Related Works}
\label{sec:context}

\subsection{Differentiation from Federated Recommendation Surveys}

Existing surveys on FRSs primarily synthesize methods based on collaborative filtering and neural networks~\cite{yang2020federated,sun2024survey,qi2025federated,zhang2025personalized}, and center on optimization strategies for communication efficiency and cryptographic protocols.
However, these works predate the widespread adoption of FMs, and treat privacy primarily as a cryptographic problem of hiding gradients.
Thus, they fail to address how pre-trained semantic priors fundamentally alter the personalization landscape of distributed recommendation. \looseness=-1

Our survey examines the integration of extensive knowledge into federated architectures. 
We move beyond simple parameter aggregation and personalization. 
Our analysis focuses on how to leverage the representational capacity of FMs individually for each client within a federated framework.
We specifically address the gap in using these models to preserve user personality for privacy-preserving recommendations, rather than just protecting raw data transmission.

\subsection{Differentiation from Foundation Model Surveys}

Conversely, recent reviews on FMs for recommendation focus predominantly on centralized paradigms~\cite{liu2023pre,wu2024survey}.
These works categorize techniques such as prompt engineering and instruction tuning under the assumption of data aggregation on a central server. 
They analyze utility maximization but largely overlook the theoretical constraints of distributed deployment, and rarely discuss the architectural implications of keeping sensitive user traits strictly local while updating the massive global backbone of FMs. \looseness=-1

Our work bridges this critical gap by providing the first structured taxonomy of Personalized Federated Foundation Models for privacy-preserving recommendations.
We emphasize the necessary \textit{architectural decoupling} required in this new era. Our survey uniquely highlights the tension between global semantic generalization and the rigorous isolation of idiosyncratic user personalities. We frame this not merely as a deployment issue, but as a fundamental shift towards active semantic preservation in recommendation systems. \looseness=-1

\section{Problem Settings}
\label{sec:bg}

This section reviews recommendation systems and the preference--personality dichotomy, the role of foundation models, and the motivation for privacy-preserving federated foundation paradigms.

\subsection{Preliminaries}

Let $\mathcal{U}$ be the set of users and $\mathcal{I}$ be the set of items.
A user $u \in \mathcal{U}$ has historical interaction data $\mathcal{D}_u$.
The full dataset of observed interactions is $\mathcal{D} = \bigcup_{u \in \mathcal{U}} \mathcal{D}_u$, where each interaction is a tuple $(u, i, r_{ui})$. 
$r_{ui}$ indicates feedback from user $u$ for item $i$.
For recommendation modeling, we use $\Theta$ to represent the learnable model parameters, comprising global parameters $\theta_g$ shared across users, and user-specific parameters $\theta_u$.
$\theta_u$ captures individual characteristics and is refined using $\mathcal{D}_u$.
In the context of FFMs, $\phi_{\mathrm{FM}}$ denotes the base FM parameters.
To adapt such FMs for specific tasks and domains, we denote global adaptable components collaboratively fine-tuned with $\phi_g$.
Correspondingly, $\phi_u$ represents user-specific adaptable components refined locally for user $u$. \looseness=-1

\subsection{Personalization in Recommendation}

\subsubsection{Recommendation Systems}
\textbf{Recommendation Systems} (RSs) aim to predict preferences $\hat{r}_{ui}$ and discover relevant items for users~\cite{ko2022survey}.
A general RS objective learns model parameters $\Theta$ by minimizing a loss $\mathcal{L}$ over $\mathcal{D}$, often with regularization $\Omega(\Theta)$~\cite{zhang2019deep}:
\begin{equation}\label{eq:obj_rs}
    \min_{\Theta} \sum_{(u,i,r_{ui}) \in \mathcal{D}} \mathcal{L}(r_{ui}, \hat{r}_{ui}(\theta_g, \theta_u)) + \Omega(\Theta).
\end{equation}
Effective personalization hinges on accurately modeling user-specific characteristics captured by $\theta_u$ in Eq.~\eqref{eq:obj_rs}~\cite{he2023survey}; however, centralized RSs aggregate all user data $\mathcal{D}_u$ for training, raising privacy concerns about the personality traits embedded in $\theta_u$~\cite{meng2018personalized}. \looseness=-1

\subsubsection{Recommendation in Federated Settings}
To mitigate such privacy risks, FL offers a distributed paradigm for collaborative refinement across users while keeping their raw data localized.
\textbf{Federated Recommendation Systems} (FRSs) apply this to maintain data locality in recommendations~\cite{sun2024survey}, as shown in Fig.~\ref{fig:CRS_vs_FRS}.
In FRSs, each user $u \in \mathcal{U}$ contributes to training with their local data $\mathcal{D}_u$.
The joint objective across clients in FRSs often learns with the shared $\theta_g$ and the user-specific $\theta_u$: \looseness=-1
\begin{equation}\label{eq:obj_frs}
    \min_{\theta_g, \{\theta_u\}_{u \in \mathcal{U}}} \sum_{u \in \mathcal{U}} 
    \frac{|\mathcal{D}_u|}{|\mathcal{D}|} 
    \sum_{(u,i,r_{ui}) \in \mathcal{D}_u}\mathcal{L}_u(r_{ui}, \hat{r}_{ui}(\theta_g,\theta_u)) + \Omega(\Theta),
\end{equation}
where $\mathcal{L}_u$ is the local loss for user $u$.
While FRSs satisfy baseline compliance requirements, the federated architecture introduces new theoretical challenges.
First, client interaction data $\mathcal{D}_u$ is often sparse and statistically heterogeneous, hindering the training of robust models that can generalize across diverse user groups~\cite{li2024federated}.
Importantly, though the server does not have access to clients' raw data, model updates themselves may still leak sensitive information under certain attack settings~\cite{chai2020secure}. \looseness=-1

\begin{figure}[t]
    \centering
    \includegraphics[width=0.75\linewidth]{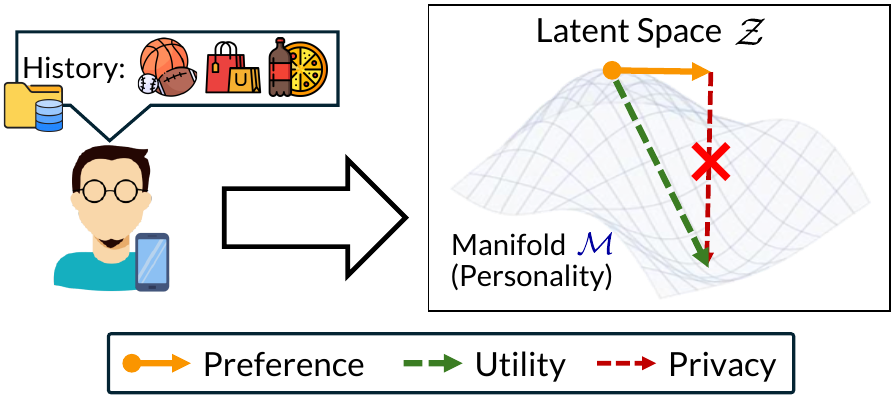}
    \caption{
    Interpretation of the dichotomy between preference and personality.
    User personality forms a stable \textbf{base manifold} $\mathcal{M}$ with the interaction history in the latent space $\mathcal{Z}$.
    Transient preferences appear as \textbf{vectors} (orange arrows) supported by this manifold.
    }
    \label{fig:manifold}
\end{figure}

\subsubsection{Dichotomy of Preference and Personality}

Effective privacy preservation of FFMs necessitates distinguishing two semantic layers within the user representation $\theta_u$ for recommendations.
We formalize this distinction through the geometry of the latent representation space $\mathcal{Z}$.
\textbf{Preference} reflects transient interests conditioned on the context.
Mathematically, it manifests as high-frequency variance in the user's interaction data $\mathcal{D}_u$.
Conversely, \textbf{Personality} functions as the invariant prior that governs the stationary distribution of the user's long-term behaviors~\cite{dhelim2022survey}.
We postulate that personality constitutes a low-dimensional \textbf{base manifold} $\mathcal{M}$ within the latent space $\mathcal{Z}$.
As shown in Fig.~\ref{fig:manifold}, preference $p_u$ functions as a directional vector learned as a perturbation originating from $\mathcal{M}$, capturing the direction of change relative to the user's long-term stable baseline. \looseness=-1

This geometric hierarchy enables mathematical decoupling.
The recommendation task in Eq.~\eqref{eq:obj_frs} primarily relies on the orientation of $p_u$ relative to item embeddings, while the exact coordinates of the manifold $\mathcal{M}$ are not strictly necessary for item scoring.
We thus designate the span of $\mathcal{M}$ as the \textit{sensitive subspace} $\mathcal{S}\subseteq\mathcal{Z}$, and treat utility and personality reconstruction as orthogonal tasks: $p_u$ is updated for accuracy while orthogonality to $\mathcal{S}$ blocks leakage of the underlying manifold structure to potential adversaries. \looseness=-1

\subsection{Foundation Models}

\textbf{Foundation Models} (FMs) represent a paradigm shift in knowledge representation.
FMs possess extensive pre-trained parameters $\phi_{\mathrm{FM}}$ trained on diverse datasets, enabling them to discern complex patterns and generate potent representations~\cite{bommasani2021opportunities}, while demonstrating robust generalization and adaptability~\cite{zhang2024scaling}.
For recommendations, adaptation involves fine-tuning a parameter set $\phi$ of the model using downstream task-specific data $\mathcal{D}'$, which is defined as:
\begin{equation} \label{eq:fm_finetune}
    \min_{\phi} \mathcal{L}(f_{\mathrm{FM}}(\phi_{\mathrm{FM}}, \phi; \mathcal{D}')) + \Psi(\phi).
\end{equation}
where $f_{\mathrm{FM}}$ is the FM, $\phi$ denotes the adaptable parameters, and $\Psi(\phi)$ is an optional regularizer.
This adaptability transfers general knowledge from $\phi_{\mathrm{FM}}$ to the recommendation context, enabling deep and semantically rich user representations even with extremely sparse user-item interactions~\cite{bommasani2021opportunities}. \looseness=-1

\begin{figure}[t]
    \centering
    \includegraphics[width=1\linewidth]{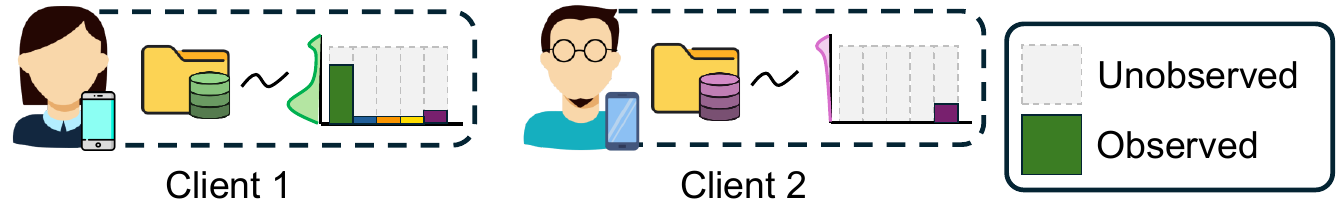}
    \caption{
    Core data challenges in FRSs.
    Each client observes only a small fraction of the items, leaving the majority unobserved locally. 
    This inherent data sparsity, combined with non-IID distributions of user behavior across clients, creates significant heterogeneity that hampers both local personalization and global model aggregation.
    }
    \label{fig:FRS_core_chall}
\end{figure}

\begin{figure*}[t]
    \centering
    \begin{subfigure}[b]{0.4\linewidth}
        \centering
        \includegraphics[width=\linewidth]{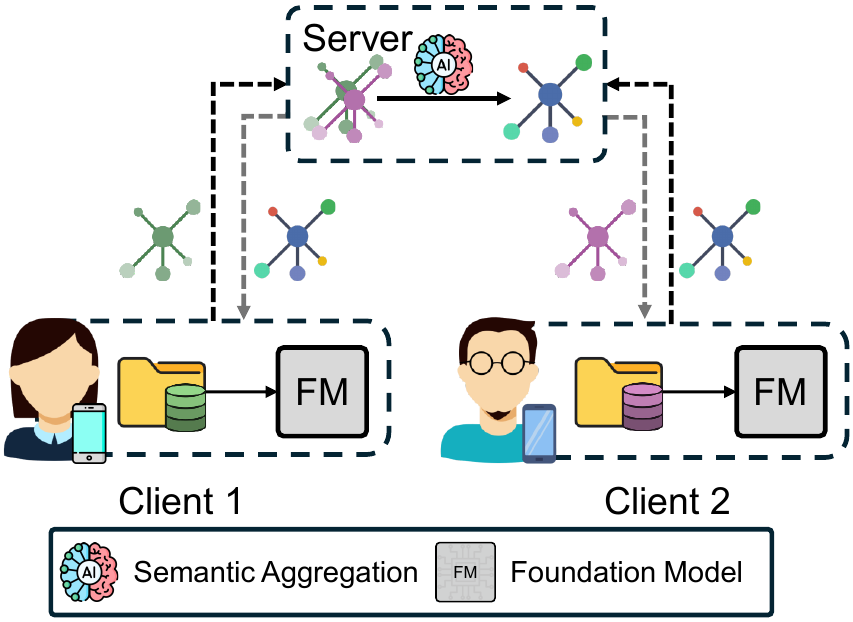}
        \caption{Federated Foundation Models}
        \label{fig:FFM}
    \end{subfigure}
    \hspace{4em}
    \begin{subfigure}[b]{0.4\linewidth}
        \centering
        \includegraphics[width=\linewidth]{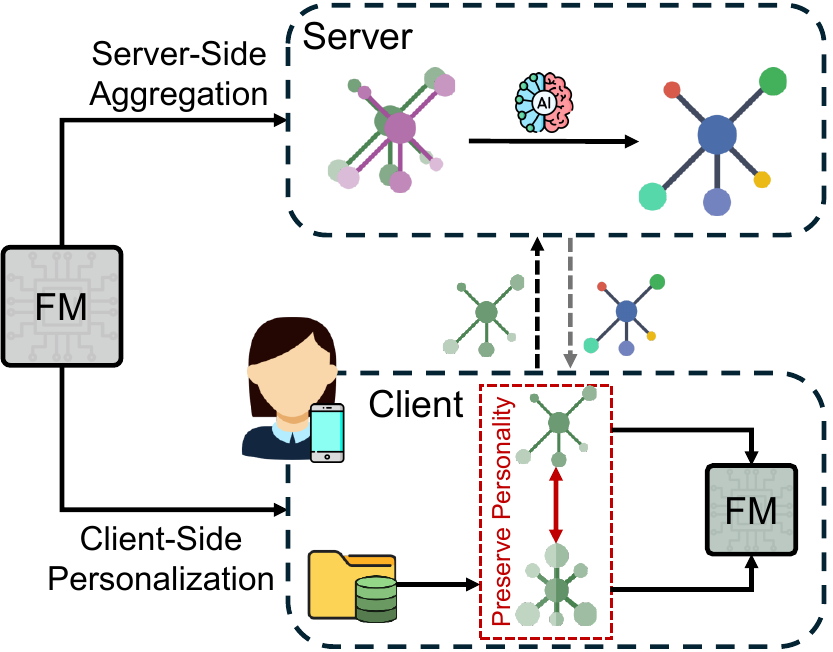}
        \caption{Personalized Federated Foundation Models}
        \label{fig:frs_fm}
    \end{subfigure}
    
    \caption{
    Evolution of the collaborative learning paradigm in recommendations.
    (a) The standard FFM approach aggregates local updates into a unified global model.
    It assimilates all learned patterns across clients without structural disentanglement, with no parameters dedicated to per-user personalization.
    (b) The proposed PFFM framework introduces active semantic defense.
    It utilizes the FMs as a shared semantic basis.
    Mechanisms such as constrained local optimization and semantic-aware aggregation architecturally decouple shared knowledge from individual personality.
    This ensures that the global model learns generalizable patterns while strictly isolating sensitive user personality.
    }
    \label{fig:paradigm_evolution}
\end{figure*}

\subsection{Federated Foundation Models}

Despite the benefits of data locality and personalized parameters $\theta_u$, FRSs still face hurdles in capturing and preserving personality.
As shown in Fig.~\ref{fig:FRS_core_chall}, local data $\mathcal{D}_u$ for each client in FRSs suffers from \textbf{Data Sparsity}, where the minimal density makes learning accurate $\theta_u$ difficult~\cite{zhang2024fedcold}.
Furthermore, \textbf{Statistical Heterogeneity} complicates deriving generalizable patterns~\cite{qi2023cross}.
Critically, FRSs do not inherently \textit{learn to preserve} personality characteristics from exposure during aggregation~\cite{chai2020secure}.
These profound limitations highlight the urgent need for more advanced and privacy-preserving modeling paradigms tailored to recommendation. \looseness=-1

FMs offer powerful representational capabilities to address this need~\cite{han2026graph}, motivating our exploration of Federated Foundation Models (FFMs) for recommendations in Fig.~\ref{fig:FFM}, in light of the FRS limitations in capturing user characteristics and protecting sensitive personality information.
We aim to harness FMs for deep personality understanding within a federated structure that safeguards these inherent user traits, necessitating a paradigm shift towards systems that actively learn to preserve the sensitive aspects of each user's personality. \looseness=-1

\section{Personalized Federated Foundation Models for Recommendation}
\label{sec:ffm_paradigm}

To realize privacy-preserving recommendations, the field is witnessing a paradigm shift toward Personalized Federated Foundation Models (PFFMs), as shown in Fig.~\ref{fig:paradigm_evolution}.
This moves the focus from passive data protection to active semantic defense~\cite{kang2025grounding}.
While standard FL ensures data locality, it inherently fails to prevent the inference of high-level attributes from model updates~\cite{ren2025advances}.
We argue that the core objective of PFFMs is to \textit{learn to preserve personality} via active semantic disentanglement: leveraging FMs to infer personality from sparse interactions while orthogonalizing sensitive traits from shared information, yielding recommendations that are both personality-aware and privacy-respectful for every user. \looseness=-1

\subsection{Foundation Model as Semantic Basis}

Unlike NLP inputs with explicit semantics, RSs process sparse behavioral signals consisting of non-semantic item identifiers, creating a fundamental semantic gap in the field.

As illustrated in Fig.~\ref{fig:frs_fm}, the PFFM paradigm posits that the FM serves as a semantic basis distributed across the federated network, establishing a continuous latent space $\mathcal{Z}$ that maps discrete interaction histories for recommendations.
Here $\mathcal{Z}$ does not merely encode item similarity but constitutes a complex behavioral manifold whose user coordinates encode intrinsic dispositions—stable personality traits that govern interaction probability distributions.

PFFMs leverage pre-trained base parameters $\phi_{\mathrm{FM}}$ of FMs to serve as a high-dimensional semantic basis, providing extensive world knowledge~\cite{ren2025advances}. 
This allows the mapping of discrete interaction history $\mathcal{D}_{u}$ into a continuous latent space $\mathcal{Z}$.
Within it, global adaptable components $\phi_g$, collaboratively learned across users, align the generic semantic basis with the recommendation domain~\cite{zhang2025multifaceted}. 
In contrast, user-specific components $\phi_u$ function as low-rank perturbations applied to the global state, encoding the stable personality traits of user $u$ which deviate from the global consensus~\cite{zhang2024transfr}.
Consequently, this process transforms into a constrained disentanglement problem that seeks to minimize a joint objective:

\begin{equation} \label{eq:ffm_global_joint_objective}
\min_{\phi_g, \{\phi_u\}} \sum_{u \in \mathcal{U}}
\frac{|\mathcal{D}_u|}{|\mathcal{D}|}
\sum_{(u,i, r_{ui})\in \mathcal{D}_u}
\mathcal{L}_{u}(r_{ui}, \hat{r}_{ui}(\phi_{\mathrm{FM}}, \phi_g, \phi_u)).
\end{equation}
The overall framework in Eq.~\eqref{eq:ffm_global_joint_objective} therefore orchestrates the federated optimization of global components and facilitates subsequent local refinement for personalization, with the challenge of maximizing utility while strictly bounding the semantic leakage inherent in $\phi_u$. \looseness=-1

\subsection{Client-Side Personalization}

Eq.~\eqref{eq:ffm_global_joint_objective} represents a dual-goal problem: representations must be maximally sufficient for utility yet minimally sufficient for personality inference, which we cast as a constrained optimization with orthogonality between shared parameters and the sensitive subspace:
\begin{equation} \label{eq:ffm_orthogonal_optimization}
\begin{aligned}
\min_{\phi_g, \{\phi_u\}}  & \quad
\sum_{u \in \mathcal{U}}
\frac{|\mathcal{D}_u|}{|\mathcal{D}|}
\sum_{(u,i, r_{ui})\in \mathcal{D}_u}
\mathcal{L}_{u}(r_{ui}, \hat{r}_{ui}(\phi_{\mathrm{FM}}, \phi_g, \phi_u)), \\
\text{s.t. } & \quad
\Psi_{\perp} (\phi_u, \phi_g) \leq \epsilon,
\forall u \in \mathcal{U}.
\end{aligned}
\end{equation}
Eq.~\eqref{eq:ffm_orthogonal_optimization} aggregates the loss over fine-grained interactions with a pivotal constraint:
$\Psi_{\perp}(\phi_u, \phi_g)$ requires the joint representation formed by $\phi_g$ and $\phi_u$ to maintain a projection onto the sensitive personality subspace $\mathcal{S}$ below a threshold $\epsilon$.
By constraining local $\phi_u$ to be orthogonal to the sensitive directions of $\phi_g$, PFFMs ensure that computed global updates remain disentangled from local user-specific profiles, transforming personalization from a soft trade-off into a hard constraint-satisfaction problem. \looseness=-1

\subsection{Server-Side Aggregation}

While local optimization in Eq.~\eqref{eq:ffm_orthogonal_optimization} mitigates leakage at the source, global robustness requires server-side intervention.
Although the server cannot access $\phi_u$, it receives global updates $\Delta \phi^{(u)}_{g}$ derived from local data~\cite{li2024federated}.
We thus redefine aggregation as semantic filtering rather than algebraic averaging, using the fixed $\phi_{\mathrm{FM}}$ as a semantic discriminator to audit incoming gradients.

We propose a semantic-aware aggregation protocol: the server inspects the alignment of $\Delta \phi^{(u)}_{g}$, and an aggregation function $g_{agg}$ projects these updates onto the known semantic basis of the FM, filtering out components highly correlated with sensitive attribute clusters.
The update rule for $\phi_g$ is:
\begin{equation}
\label{eq:ffm_semantic_filter}
\phi_g^{t+1} \leftarrow \phi_g^t - \eta \cdot G_{\perp} \left(
\{\Delta \phi^{(u),t+1}_{g}\}_{u \in \mathcal{U}}, \{\zeta_u^t\}_{u \in \mathcal{U}}; \phi_{\mathrm{FM}}
\right).
\end{equation}
Here $\Delta \phi^{(u),t+1}_{g}$ is the global-parameter update from client $u$, and $\zeta_u^t$ denotes distributional statistics uploaded to assist semantic analysis.
The operator $G_{\perp}$ uses $\phi_{\mathrm{FM}}$ to analyze $\zeta_u^t$, identifies sensitive semantic directions in the update batch, and projects out the corresponding components of $\Delta\phi_g^{(u)}$ before aggregation.
This guarantees that $\phi_g$ assimilates generalizable patterns while rejecting personality fingerprints carried by gradients, jointly constituting---with Eq.~\eqref{eq:ffm_orthogonal_optimization}---our active-preservation paradigm.

\subsection{Learning to Preserve Personality}

The proposed framework establishes a fundamental distinction from standard FRSs, which lies in the controllability of the latent representation space.
Traditional FRSs train user representations in the subspace $\mathcal{S}$ of personality traits from random initialization. 
Without a prior basis, the resulting latent space $\mathcal{Z}$ is isotropic and rotationally invariant~\cite{maiorca2023latent}.
In such a manifold, the basis of $\mathcal{S}$ is unknown and time-varying, and the gradient update vector $\Delta \theta$ derived from local data represents a coupled signal.
Consequently, the projection of the update onto $\mathcal{S}$ is undefined, and separating generic preferences from sensitive personality is mathematically intractable---structural isolation alone therefore fails to prevent this semantic leakage. \looseness=-1

PFFMs advocate a paradigm shift from structural to semantic privacy boundaries, predicated on a fixed semantic reference frame uniquely provided by the pre-trained FFM.
This prior breaks the rotational invariance and allows for the rigorous definition of $\mathcal{S}$ within $\mathcal{Z}$, so the privacy boundary in recommendations is reformulated as:
\begin{equation}
\label{eq:privacy_boundary_condition} 
\langle \Delta \phi_{g}, \zeta_u \rangle \to 0, \quad \forall u \in \mathcal{U}. 
\end{equation}
Even if $\Delta\phi_{g}$ is intercepted, its projection onto the semantic directions of personality is theoretically nullified, shifting the privacy boundary from an architectural heuristic to a verifiable geometric constraint and preserving privacy both by where the data resides and by how the representation is constructed. \looseness=-1

\subsection{Taxonomy of Instantiating the Framework}
\label{subsec:taxonomy}

To map existing research into the above framework, we categorize PFFMs based on the specific architectural realization of the decoupled parameters $\phi_u$ and $\phi_g$~\cite{ren2025advances}.

\paragraph{Adapter-based Disentanglement.}
In this category, $\phi_g$ constitutes the frozen backbone of the FM plus shared adapter modules~\cite{zhang2025multifaceted}.
The user-specific $\phi_u$ is instantiated as local lightweight adapters injected into transformer layers~\cite{sun2024exploring}, structurally enforcing $\Psi_{\perp}$ via separate matrices.
Privacy is preserved by keeping local LoRA matrices on-device while only shared adapters or statistical prototypes are aggregated~\cite{yang2024dual}, satisfying Eq.~\eqref{eq:privacy_boundary_condition} by construction.

\paragraph{Prompt-based Projection.}
Here the FM $\phi_{\mathrm{FM}}$ remains static.
The personalization $\phi_u$ is modeled as a set of continuous learnable vectors prepended to the input embeddings~\cite{liu2023pre}.
These prompts act as a geometric transformation function $f: \mathcal{X} \to \mathcal{Z}$.
It projects user behavior into the semantic space of the FM~\cite{li2026federated,su2024federated}.
Since prompts are detached from the model weights, they offer a natural boundary for the semantic isolation described in Eq.~\eqref{eq:privacy_boundary_condition}~\cite{luo2025mixture}.

\paragraph{Instruction-tuned Alignment.}
This stream focuses on semantic alignment of the global model $\phi_g$ through local instruction tuning on $\mathcal{D}_u$~\cite{zhuang2023foundation}.
Here $\phi_u$ is the local-update offset diverging from the global gradient~\cite{wang2026federated}; since $\phi_u$ and $\phi_g$ share physical parameters, advanced aggregation in Eq.~\eqref{eq:ffm_semantic_filter} is critical for filtering personality-sensitive gradients~\cite{wang2025fedsc}.

\section{Open Problems}
\label{sec:challenges}

\begin{figure}[t]
    \centering
    \includegraphics[width=1\linewidth]{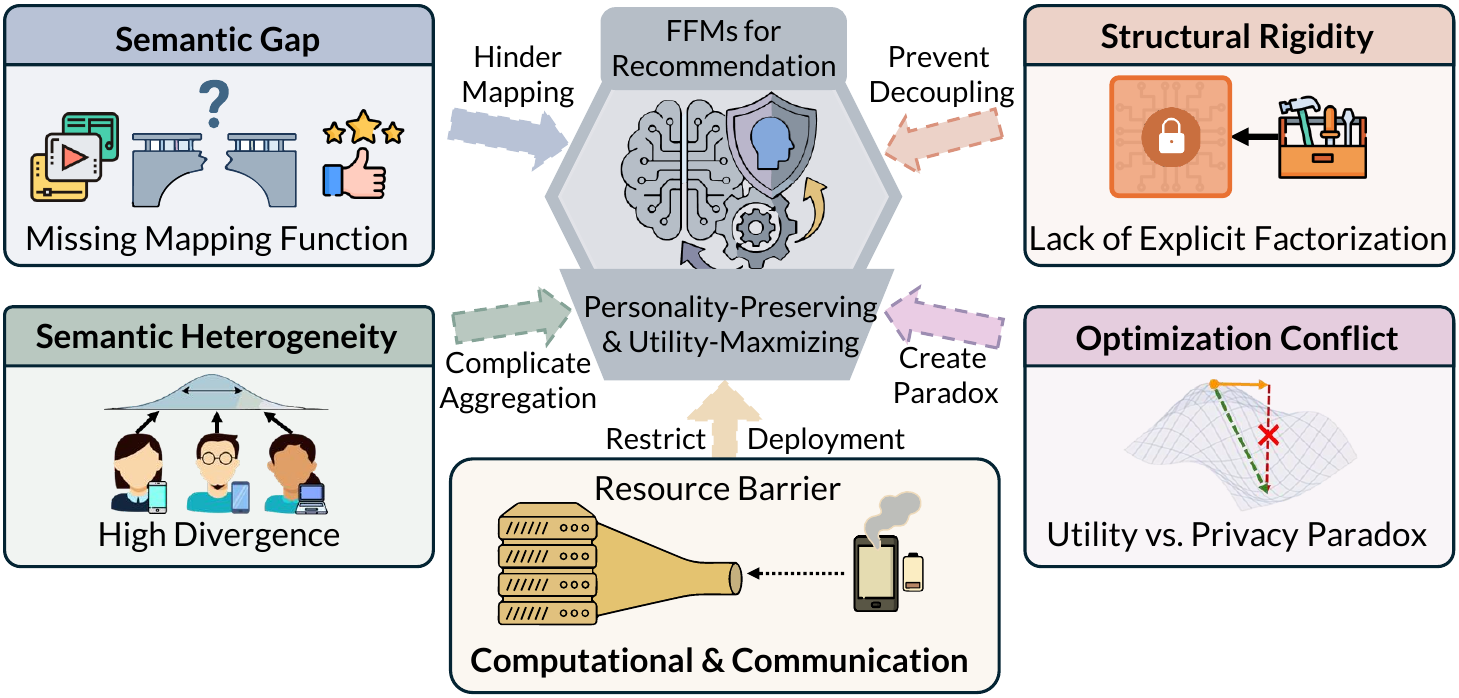}
    \caption{
    Impediments to realizing Personalized Federated Foundation Models for privacy-preserving Recommendations. 
    }
    \label{fig:problems}
\end{figure}

Realizing FFMs that learn to preserve personality involves overcoming fundamental contradictions in model optimization and data representation.
We identify five core problems, illustrated in Fig.~\ref{fig:problems}, that hinder current progress for recommendations.

\subsection{Semantic Gap}

A primary theoretical hurdle is the dissonance between observed data and the latent ground-truth personality $p_u$.
In the FFM setting, $\phi_u$ is optimized to maximize the likelihood of observed interactions $\mathcal{D}_u$ as a surrogate for recovering $p_u$:
\begin{equation}
\label{eq:max_likelihood}
\max_{\phi_u} \; P(\mathcal{D}_u \mid \phi_u; \phi_{\mathrm{FM}}, \phi_g).
\end{equation}
However, $\mathcal{D}_u$ is sparse and noisy, often reflecting transient interests rather than stable traits $p_u$.
The core challenge is the absence of a verified mapping $f: \mathcal{Z} \to \mathcal{P}$ from the learned latent space $\mathcal{Z}$ to the psychological space $\mathcal{P}$.
Without ground-truth labels for $p_u$, the model relies on spurious correlations~\cite{dhelim2022survey}, yielding a $\phi_u$ where transient noise is indistinguishable from stable personality features, so maximizing Eq.~\eqref{eq:max_likelihood} does not guarantee that $\phi_u$ accurately encodes $p_u$, creating a persistent semantic gap. \looseness=-1


\subsection{Structural Rigidity}

Current architectures lack the structural inductive bias required for recommendations, especially in the federated context~\cite{zhang2024federated}.
Most FMs largely repurpose models pre-trained on centralized textual corpora to minimize sequence entropy, whereas recommendations rely on collaborative signals to minimize $\mathcal{L}_u$.
Ideally, model parameters should be decomposable into a shared knowledge module and a private personality module:
\begin{equation}
\label{eq:decouple_fm}
\phi = \phi_g \oplus \phi_u, \quad \text{s.t. } \phi_{g} \perp \phi_{u}.
\end{equation}
However, $\phi_{\mathrm{FM}}$ is monolithic and does not admit the explicit factorization in Eq.~\eqref{eq:decouple_fm}~\cite{yang2024dual}; forcefully adapting text-centric models to this decoupled structure incurs suboptimal performance and high computational costs, preventing physical isolation of sensitive personality from the shared global model~\cite{melis2019exploiting,zhu2019deep}. \looseness=-1

\subsection{Semantic Heterogeneity}

FFMs face a unique form of heterogeneity, which we term \textit{Semantic Heterogeneity}, distinct from statistical non-IIDness in traditional FL~\cite{lin2022fednlp}.
It arises from the varying degrees of alignment between the FM's pre-trained prior knowledge $P_{\mathrm{FM}}$ and the local behavioral distribution $P_u$ of different users~\cite{wang2025fedsc}.
Some users align well with the FM's generalized knowledge, while others exhibit idiosyncratic patterns that contradict its priors.
We formalize this as the variance of semantic divergence across $\mathcal{U}$:
\begin{equation}
\label{eq:semantic_heterogeneity}
\sigma^2 = \text{Var}_{u \in \mathcal{U}} \left( D_{KL}(P_u(\mathbf{z}) \| P_{\mathrm{FM}}(\mathbf{z})) \right).
\end{equation}
A high $\sigma^2$ implies that a single global $\phi_g$ cannot accommodate all users: forcing alignment with $\phi_g$ dilutes the personality of high-divergence users, while excessive local deviation $\phi_u$ causes catastrophic forgetting of the FM's general capabilities, creating a fundamental dilemma in selecting the appropriate aggregation weight for each user. \looseness=-1

\subsection{Optimization Conflict}

FFMs face an inherent paradox between utility maximization and privacy constraints~\cite{zhang2023trading}.
As discussed in Sec.~\ref{sec:ffm_paradigm}, preserving personality requires gradient updates orthogonal to the sensitive subspace $\mathcal{S}$, i.e., the local gradient $\nabla \mathcal{L}_u$ must satisfy:
\begin{equation}
\label{eq:proj}
\| \text{Proj}_{\mathcal{S}} (\nabla \mathcal{L}_u) \| = 0.
\end{equation}
Eq.~\eqref{eq:proj} means the projection of $\nabla \mathcal{L}_u$ onto $\mathcal{S}$ is zero; however, utility maximization relies on leveraging personality traits $p_u$ to predict behavior, so if $\nabla \mathcal{L}_u$ provides useful information for personalization, it implies a correlation:
\begin{equation}
| \text{Corr}(\nabla \mathcal{L}_u, p_u) | > 0.
\end{equation}
Since $p_u$ lies within $\mathcal{S}$, the two conditions are contradictory; standard gradient descent cannot satisfy both simultaneously without degrading utility, leading to an inevitable optimization conflict where the pursuit of accuracy inherently compromises the semantic orthogonality required for privacy. \looseness=-1

\subsection{Computational and Communication Overhead}

Beyond theoretical paradoxes, deploying FFMs faces rigid physical constraints.
Clients in federated settings typically possess limited computational capacity and battery life~\cite{yang2019federated}.
However, FM inference and fine-tuning require substantial FLOPs and memory bandwidth~\cite{sun2024exploring}; even with Parameter-Efficient Fine-Tuning (PEFT), the forward pass still loads the massive $\phi_{\mathrm{FM}}$ into memory~\cite{yang2024dual}, creating a resource barrier.
Communication bandwidth further restricts the frequency and dimension of $\Delta\phi_g$ transmitted between clients and server~\cite{vu2024analysis}, conflicting with the depth of semantic reasoning required for personality analysis; standard compression~\cite{dantas2024comprehensive} often degrades the ability to capture subtle personality nuances during transmission. \looseness=-1

\section{Future Directions}
\label{sec:future_directions}

\begin{figure}[t]
    \centering
    \includegraphics[width=.9\linewidth]{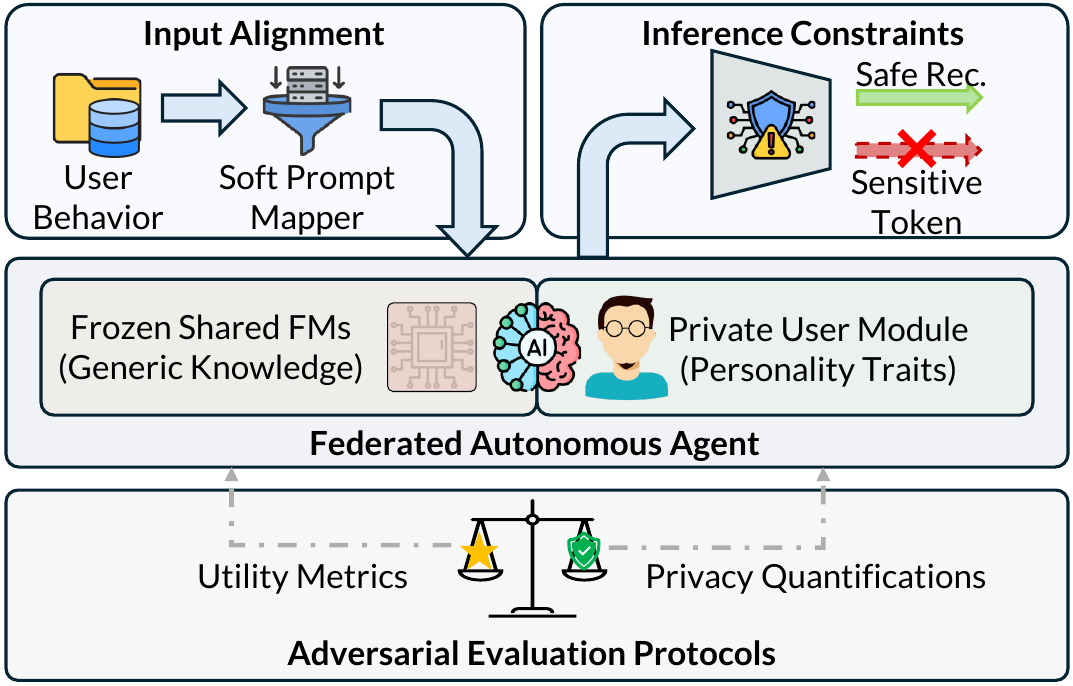}
    \caption{
    Framework for future directions in personality-preserving Federated Foundation Models for Recommendations.
    }
    \label{fig:directions}
\end{figure}

The formal challenges in Sec.~\ref{sec:challenges} suggest that incremental improvements are insufficient; we advocate new research paradigms in the federated context.
As shown in Fig.~\ref{fig:directions}, these directions explicitly address the mathematical contradictions above, aiming for personality-aware recommendations while actively \textit{learning to preserve} the sensitive aspects of user personality. \looseness=-1

\subsection{Federated Autonomous Agents}

The semantic gap between transient feedback and stable personality arises from the limitations of single-step prediction.
\textbf{Federated Autonomous Agents} offer a viable solution by reformulating the interaction paradigm~\cite{deng2025fedslate}.
We advocate treating the FFM as a local agent with policy $\pi$, modeling user interactions as a Markov decision process~\cite{lian2024recai}.
Rather than maximizing feedback likelihood, the agent optimizes a state-value function $V_{\pi}(s_t)$:
\begin{equation}
\label{eq:agent_objective}
V_{\pi}(s_t) = \mathbb{E} \left[ 
\sum_{k=0}^{\infty} \gamma^k R(s_{t+k}, a_{t+k} \mid s_t, p_{u} )
\right].
\end{equation}
Here, $\gamma$ is a discount factor and $s_t$ explicitly includes the latent personality variable $p_{u}$, while $R(\cdot)$ enforces consistency in the agent's behavior $a_{t}$ over $t$, so the agent learns to encode stable personality traits as necessary components for coherent long-term planning, shifting from fitting noisy logs to learning a coherent policy $\pi(a|s_t, p_{u})$.
Recent agent-based recommenders, including multi-step planners and tool-using LLM agents~\cite{lian2024recai}, demonstrate this trajectory; their federated extensions must additionally guarantee that $p_{u}$ never crosses the local boundary, e.g., by replacing remote rollouts with locally simulated trajectories that disclose only utility statistics. \looseness=-1

\subsection{Parameter-Efficient Modular Design}

To overcome the structural rigidity of monolithic models, research must adopt Parameter-Efficient Modular Design that physically decouples general knowledge from user-specific traits~\cite{zhang2024federated}.
We advocate treating personality parameters as independent lightweight modules such as adapters or hyper-networks~\cite{yang2024dual}: the massive $\phi_{\mathrm{FM}}$ remains frozen and shared while only $\{\phi_u\}$ are updated locally.
Promising instantiations include per-user low-rank LoRA factors, hyper-networks that generate $\phi_u$ conditioned on user embeddings, and modular routing schemes that isolate personality experts from the shared backbone, all of which keep the gradient dynamics of personality strictly separated from $\phi_{\mathrm{FM}}$ and thus enable rigorous encryption or differential privacy on $\phi_u$ alone without degrading FM utility. \looseness=-1

\subsection{Personalized Prompt Learning}

Semantic heterogeneity across users motivates \textbf{Personalized Prompt Learning}: instead of modifying weights, the system learns user-specific soft prompts that act as continuous mappings~\cite{luo2025mixture}, projecting each user's behavioral manifold into the foundation model's semantic space~\cite{li2024visual}.
Designs such as mixture-of-prompts gating, hierarchical prompt prototypes shared across similar users, and meta-prompt initialization for fast local adaptation~\cite{su2024federated} each trade off expressivity, communication cost, and personality privacy differently.
Minimizing the alignment error locally through prompts reduces the heterogeneity of uploaded representations and allows the global model to aggregate insights effectively while respecting the distinct semantic topology of each personality. \looseness=-1

\subsection{Inference-Time Privacy Constraints}

The optimization conflict between utility and privacy is hard to resolve during training due to collinear gradients~\cite{wang2025privacy}.
\textbf{Inference-Time Privacy Constraints} offer a pragmatic alternative by shifting the defense to the decoding stage: in generative recommendation, the system enforces privacy without altering parameters~\cite{luo2025prompt} by integrating constraints into beam search or sampling, suppressing tokens with high mutual information w.r.t.\ sensitive attributes.
Concrete instantiations include logit reweighting against sensitive token clusters, constrained nucleus sampling that masks personality-revealing continuations, and gradient-free differential privacy injected at the decoder, so that outputs remain useful while statistically invariant to private personality traits.
Compared with training-time defenses, this strategy decouples privacy budgets from optimization trajectories and is therefore robust to subsequent fine-tuning of the shared model. \looseness=-1

\subsection{Adversarial Evaluation Protocols}

The lack of standardized metrics hinders the verification of personality-preserving FFMs, motivating rigorous \textbf{Adversarial Evaluation Protocols}.
We advocate a dual-metric system: utility is measured by standard ranking metrics, while privacy is quantified via \textit{Personality Inference Attacks} in which an adversarial model reconstructs the user's base-manifold coordinates from exposed model updates or outputs~\cite{zhang2023comprehensive,chang2024efficient}.
Privacy success is then scored by the inverse of attacker accuracy or the reduction in mutual information, which a robust system must minimize while maximizing recommendation utility.
Building reproducible benchmarks further requires pairing standard recommendation corpora with personality labels obtained from Big-Five inventories or LLM-based persona elicitation, so that the trade-off frontier between utility and personality leakage can be tracked across methods.
Robust evaluation must cover both gradient-side leakage during training and output-side inference at deployment, since these expose distinct facets of personality and warrant distinct defenses. \looseness=-1

\section{Conclusion}
\label{sec:conclusion}

This survey provides a comprehensive analysis of Personalized Federated Foundation Models for privacy-preserving recommendation, examining how the architecture resolves the conflict between generalizable knowledge and personalized adaptation within federated settings.
A key insight is the necessity of shifting from passive data protection to active semantic preservation: foundation models can infer nuanced personality traits, but decentralized protocols must structurally isolate these attributes from shared parameters to ensure privacy.
Advancing this paradigm requires rigorous mechanisms that maintain semantic boundaries around user identity, ensuring trustworthy recommendation systems that protect both user privacy and personality.
We hope the taxonomy of architectural realizations and the geometric privacy boundary articulated here serve as a foundation for next-generation federated recommendation systems that genuinely empower users with both relevance and self-determination over their digital identity, and that the open problems we identify catalyze cross-disciplinary work bridging federated optimization, foundation-model adaptation, and privacy-aware semantic geometry. \looseness=-1

\bibliographystyle{named}
\bibliography{ijcai26}

\end{document}